\documentclass[runningheads]{llncs}
\usepackage{graphicx}

\usepackage{lineno,hyperref,amsfonts,algorithm,algorithmic}

\usepackage{amssymb,graphics,subfigure,graphicx,subfigure,verbatim}
\usepackage{amsmath}
\usepackage{floatflt}
\usepackage{indentfirst}
\usepackage{latexsym,bm}
\usepackage{multicol,multirow,booktabs}
\usepackage{rotating}
\usepackage{lineno}
\usepackage{color}
\begin{document}
\title{Graph Neural Networks for Unsupervised Domain Adaptation of Histopathological Image Analytics}
\titlerunning{Graph Neural Networks for Unsupervised Domain Adaptation}
%


\author{Dou Xu$^1$, Chang Cai$^1$, Chaowei Fang$^2$, Bin Kong$^3$, Jihua Zhu$^1$, Zhongyu Li$^{1,*}$}
\institute{$^1$School of Software Engineering, Xi’an Jiaotong University, Xi’an, China\\
$^2$School of Artificial Intelligence, Xidian University, Xi’an, China\\
$^3$Department of Computer Science, University of North Carolina at Charlotte, Charlotte, USA\\
\email{zhongyuli@xjtu.edu.cn}}

\maketitle              
\begin{abstract}
Annotating histopathological images is a time-consuming and labor-intensive process, which requires broad-certificated pathologists carefully examining large-scale whole-slide images from cells to tissues. Recent frontiers of transfer learning techniques have been widely investigated for image understanding tasks with limited annotations. However, when applied for the analytics of histology images, few of them can effectively avoid the performance degradation caused by the domain discrepancy between the source training dataset and the target dataset, such as different tissues, staining appearances, and imaging devices. To this end, we present a novel method for the unsupervised domain adaptation in histopathological image analysis, based on a backbone for embedding input images into a feature space, and a graph neural layer for propagating the supervision signals of images with labels. The graph model is set up by connecting every image with its close neighbors in the embedded feature space. Then graph neural network is employed to synthesize new feature representation from every image. During the training stage, target samples with confident inferences are dynamically allocated with pseudo labels. The cross-entropy loss function is used to constrain the predictions of source samples with manually marked labels and target samples with pseudo labels. Furthermore, the maximum mean diversity is adopted to facilitate the extraction of domain-invariant feature representations, and contrastive learning is exploited to enhance the category discrimination of learned features. In experiments of the unsupervised domain adaptation for histopathological image classification, our method achieves state-of-the-art performance on four public datasets.

\keywords{Histopathological Image Analysis \and Domain Adaptation \and Graph Neural Networks \and Computer-Aided Diagnosis.}
\end{abstract}


\section{Introduction}
Histopathological images have been widely applied in diagnosing and screening many kinds of cancers. However, large-scale whole-slide images (WSIs) pose significant challenges for computer-aided diagnosis (CAD) systems which usually rely on training machine learning models, e.g., deep convolutional neural networks, with a large number of well-annotated samples. To collect the training dataset, broad-certificated pathologists need to carefully examine hundreds to thousands of images with gigabyte size. The annotation process is very time-consuming and highly labor-intensive. 

Several domain adaptation methods have been developed for specific tasks in histopathological image analytics. For example, Lafarge \emph{et al}.~\cite{Lafarge2017} first proposed a systematic solution based on domain-adversarial neural networks (DANN)~\cite{Ganin2016}, to address the appearance variabilities of breast cancer histopathology images for mitosis detection. Subsequently, Wollmann \emph{et al}.~\cite{Wollmann2018} exploited a Cycle-Consistent Generative Adversarial Network (CycleGAN) to develop a domain adaptation method for classifying whole-slide images and grading the patient level of the breast cancer. Aiming at the Gleason grading across two prostate cancer datasets, Ren \emph{et al}.~\cite{Ren2018} devised a domain adaptation framework  which is driven by the adversarial learning on the basis of the Siamese architecture. More recently, Zhang \emph{et al}.~\cite{Zhang2019} put forward a deep unsupervised domain adaptation method to adapt deep models trained on the labeled whole-slide image domain to the unlabeled microscopy image domain. It was implemented by reducing domain discrepancies via adversarial learning and entropy minimization, and training samples were reweighted to alleviate class imbalances.

Though these methods can effectively relieve the influence of the domain shift issue in specific histopathological CAD tasks, it is still challenging for them to extend well for more general tasks in histopathological image analytics. First of all, histopathological images usually showcase various appearances, because of different staining methods (e.g., H\&E, Masson's trichrome), tissues (e.g., breast and prostate), and imaging devices (e.g., light and electron microscope). In comparison with commonly investigated cases, discrepancies among histopathological image datasets pose significant challenges for the domain adaptation problem. Secondly, the large intra-class variations and small inter-class variations in histopathological images make it difficult to identify fine-grained disease categories. Thirdly, existing domain adaptation methods devised for histopathological image analytics are mainly based on the framework of the antagonistic learning. The objective function is approximately solved through minimax optimization algorithms, which are inefficient and frequently cause model collapse. As far as we know, none of existing methods in this field takes the relationships between similar pathological tissues into consideration.

Taking the above challenges into account, we propose a generalized framework for the unsupervised domain adaptation in histopathological image analysis. Firstly, a backbone CNN is used to extract deep features from input images and the maximum mean diversity is used to transform the feature distributions of source images and target images into a uniform feature space. For a target image with a confident prediction, we directly regard the class with the maximum probability as its pseudo~\cite{pseudolabel} label. Samples with small distances in the feature space are highly likely to share the same semantic category. Motivated by this point, we devise a graph model, in which samples with relatively close distances in the feature space are linked, to hallucinate new feature representations via mixing up every anchor sample and its neighboring samples. The categories of these new feature representations are assumed to inherit from their anchor samples. The proposed graph model can make full usage of unlabeled target images and is beneficial for  increasing the generalization ability of the deep model according to mixup based methods~\cite{zhang2017mixup,berthelot2019mixmatch}. 
Finally, aiming at increasing the discriminative capability of the learned features, a contrastive loss function is used to constrain the distances between samples having the same category within a small value while preserving large values for distances between samples from different categories.
The experimental results show that the proposed method achieves excellent performance on multiple histopathological image classification tasks under the scenario of unsupervised domain adaptation.

\section{Methodology}
\begin{figure}[t]
\centering
\includegraphics[width=1\linewidth]{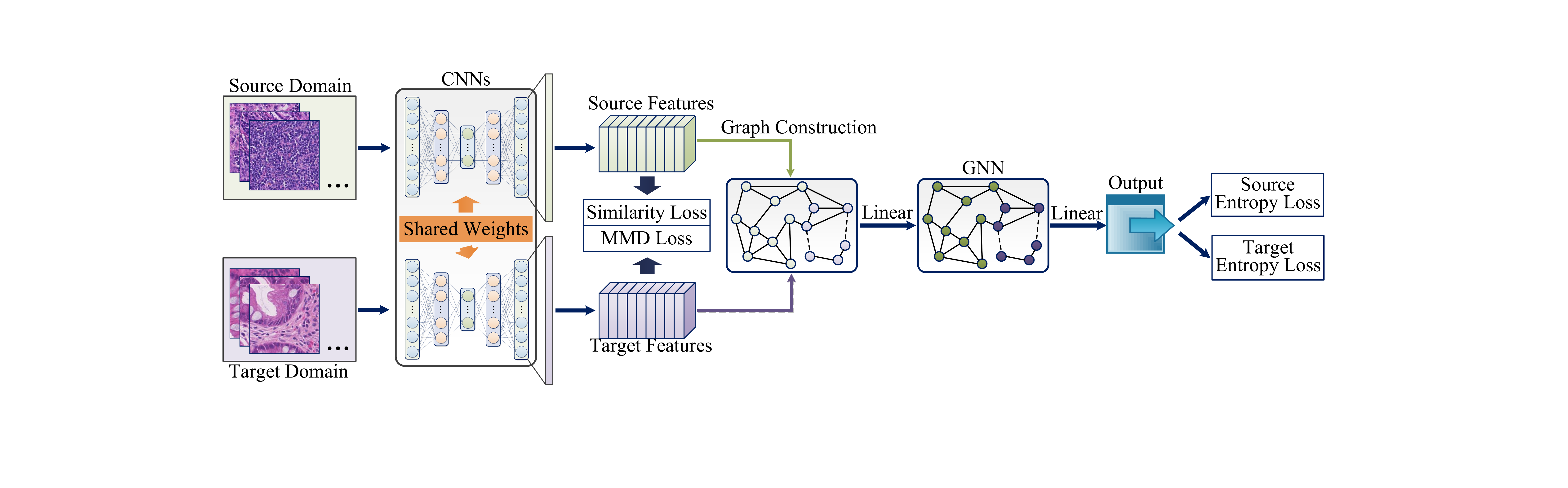}
\caption{Overview of our proposed framework for the unsupervised domain adaptation of histopathological image analytics.}
\label{fig: overview}
\end{figure}
\noindent\textbf{Overview:} 
Fig.~\ref{fig: overview} presents an overview of the proposed framework for the unsupervised domain adaptation of histopathological image analytics. Particularly, the image set  $\mathbb X^s=\{x^s_i|i=1,\cdots,N^s\}$ in source domain is provided with well-annotated labels, while the image set $\mathbb X^t=\{x^t_j|j=1,\cdots,N^t\}$ in the target domain has no labels. Let $l_i^s\in\{1,\cdots,m\}$ be the ground-truth label of $x^s_i$. Here, $m$ represents the number of classes.  Given these datasets, we can train the model with images in source domain, and the goal is to adapt the trained CNN to fit in with the target domain.  The Maximum Mean Discrepancy (MMD) loss is used to reduce the inter-domain discrepancy between feature representations of source and target images. Moreover, for the purpose of making full usage of unlabeled target images, we dynamically assign pseudo labels~\cite{pseudolabel} to target images with confident predictions. A GNN~\cite{graphconv2018} module is devised to propagate supervision signals from images with manual/pseudo~\cite{pseudolabel} annotations to their close neighbors in the feature space. Additionally, the contrastive feature constraint is employed to improve the discriminativeness of the learned feature representations.

\noindent\textbf{Reducing Domain Disparity:} Firstly, the backbone CNN $\phi$ is used to simultaneously extract features from the source and target domain images. Due to the domain discrepancy, the distributions of the extracted source and target image features are totally different even if their labels are the same. To reduce the domain discrepancy, we use the Maximum Mean Discrepancy (MMD) loss~\cite{venkateswara2017deep} to enforce $\phi$ to extract similar features for source and target domain images:
\begin{align}  \label{eq:loss-mmd}
L_{mmd} = \left\| \mathbb{E}[\phi(x_i^s)] - \mathbb{E}[\phi(x_j^t)] \right\|_{{H_k}}^2
\end{align}
where ${H_k}$ denotes a reproducing kernel Hilbert space associated with the nonlinear mapping $\phi ( \cdot )$  and a kernel $k(x_i,x_j) = \left\langle {\phi (x_i),\phi (x_j)} \right\rangle$. The feature kernel $k( \cdot )$  is determined as a convex
combination of $\kappa$ PSD kernels, $\{ {k_m}\} _{m = 1}^\kappa ,K : = \{ k:k = \sum\limits_{m = 1}^\kappa  {{\beta _m}{k_m},\sum\limits_{m = 1}^\kappa  {{\beta _m} = 1,} } {\beta _m} \ge 0,\forall m\}$.

Intuitively, the representations of two images should be the similar if they have the same label. Nevertheless, only applying MMD loss is not enough to enforcing this property. To address this issue, we propose the feature similarity loss to enforce the images with the same label to have similar features~\cite{featureLoss2018}:
\begin{equation} \label{eq:loss-feat}
L_g= \sum_{j>i,l_i\neq-1,l_j\neq -1}\mathbf 1(l_i=l_j) {\left\| {\phi(x_i) - \phi(x_j)} \right\|^2} + \\
\mathbf 1(l_i\neq l_j) \max {(0,d - \left\|\phi(x_i) - \phi(x_j) \right\|^2)}
\end{equation}
where $d$ is a pre-defined margin,set to 2.  With the help of the above loss function, intra-class variations can be suppressed while inter-class distances are kept larger than the pre-defined margin.

\noindent\textbf{Graph Neural Networks Module:} We use a GNN module $\psi$ to further enforce the images with the same label to have similar feature representations. In this work, Higher-order Graph Neural Networks~\cite{graphconv2018} is used to consider the higher-order connection between graph nodes. More specifically, as shown in Fig.~\ref{fig: overview}, for each mini-batch of source and target domain images, we first use the extracted features to construct a graph, with each vertex denoting features from a source/target domain image. An edge exists between two pair vertices only if the distance between the corresponding image features is less than a threshold $T$. The image representations are then fed into a fully connected layer and a GNN~\cite{graphconv2018} layer $\psi$ to smooth the representations to force similar image representation in the feature space. The updated representation $f(x_i)$ are calculated as follows:
\begin{equation}
f(x_i)={\theta _1}{\textrm{ReLU}(\phi(x_i))w} + \sum\limits_{j \in \mathcal N(i)} {{\theta _2}{\textrm{ReLU}(\phi(x_j))w}}
\end{equation}
where $\theta_1$ and $\theta_2$ are learnable variables in the GNN~\cite{graphconv2018} layer. $x_i$ is the source/target domain image. $\mathcal N(i)$ represents the set of neighbors of the $i^{th}$ node. $w$ indicates the parameter of the linear layer. The updated feature is followed by another fully-connected layer and a softmax layer, yielding the final prediction $\hat y_i$ for each node.

\noindent\textbf{Pseudo Annotations of Target Images:} Pseudo-labeling~\cite{pseudolabel} has been widely employed to address semi-supervised learning problems. In this work, we leverage pseudo-labeling~\cite{pseudolabel} to annotate the unlabeled target domain images. Specifically, if the prediction for the target domain image $x^t_j$ is above a threshold $\epsilon$, we use its prediction (the pseudo label~\cite{pseudolabel}) to supervise the training of the proposed network.
\begin{equation}
l_i^t=\begin{cases}
j^\star & j^\star=\arg\max_j y_i^t(j),\; \textrm{and}\; y_i^t(j^\star)>\epsilon\\
-1 & \textrm{otherwise}
\end{cases}
\end{equation}
where $y_i^t$ is $x_i^t$ which is obtained by two-layer full connection transformation and $y_i^t(j)$ is the $j$-th element of $y_i^t$, and $\epsilon$($=0.97$) is a constant. $l_i^t=-1$ indicates the label of $x_i^t$ is remained unknown yet. $y_i^t$ is produced with the online model and varies as the optimization process continues.

\noindent\textbf{Loss Function:} In summary, the total loss to train the proposed framework is as follows:
\begin{align}
 L_{total} = L_{mmd} + {L_g} + {L_{ce}}
\end{align}
where $L_{mmd}$ and ${L_g}$ are the MMD and graph loss, which are detailed in Eq.~\ref{eq:loss-mmd} and~\ref{eq:loss-feat}. $L_{ce}$ is the node-level cross-entropy loss, which is applied to each of the source domain image nodes and the target domain image nodes with pseudo labels~\cite{pseudolabel}.
As each of the modules is fully differentiable, the whole network can be trained in end-to-end to transfer the classification model of the source histopathological image dataset into the target domain.

\section{Experiment}
\noindent\textbf{Experimental Settings:} 
We validate the proposed unsupervised domain adaptation algorithm for the analytics of histopathological images on four public data sets, i.e., CAMELYON16~\cite{Bejnordi2017} (abbreviated as CAM16), BreakHis~\cite{Spanhol2016}, GlaS~\cite{Warwick2015}, and DigestPath~\cite{Digest2019}. These datasets were originally released for different diagnostic applications, including the detection of breast cancer metastases, glandular structures, colonoscopy tissue, etc, which demonstrate diversified appearances. In our experiment, we crop out positive and negative patches (i.e., whether the patches contain tumor regions) with the size of $224 \times 224$ from these datasets for training and testing. For CAMELYON16~\cite{Bejnordi2017}, the positive patches come from the annotated foreground region, and the negative patches come from the background region in the whole slide images. For BreakHis~\cite{Spanhol2016}, the set of malignant images under 40X magnification of breakhis are regarded as positive samples, while the negative patches are formed by those benign images.
For DigestPath~\cite{Digest2019}, the positive patches come from 77 image areas of 20 WSI, which are marked with bounding boxes. 378 image are sampled from 79 WSI as negative patches. The patches are all under 40X magnification.
For GlaS~\cite{Warwick2015}, the positive patches come from 37 images. The negative set contains 48 images. The patches are all under 20X magnification. Table.~\ref{datasets} presents the number of patches of both positive and negative samples. 

All ground truth annotations of the images in target domain are only used for evaluation and will not be used in the training phase. Image patches are normalized to zero mean and unit variance. To train our model, we use images of the same size and enhance them with rotation, scaling, and affine transformations to reduce over-fitting. We implement the proposed framework with Pytorch. The backbone CNN used for feature extraction is based on ResNet18~\cite{resnet}. 1 graph convolution layer is attached to the backbone CNN. 2 fully connected layers are used to predict the final result from the deep features. Feature space mapping threshold is set as 150. We use Adam optimizer with the batch size of 256, in which the source data and target data are 128 respectively. Each batchsize source data and target data are randomly selected. The learning rate is set as $0.001$. The weight decay is set as $10^{-6}$. During the inference phase, the graph convolution layer is removed, and the two linear layers are directly used to estimate the final prediction from the feature of the input image. 
\begin{table*}[t]
\centering
\caption{Number of positive and negative samples we extracted from the four  public datasets.}
\begin{tabular}{l|c|c|c|c}
\hline
         & CAMELYON16~\cite{Bejnordi2017} & BreakHis~\cite{Spanhol2016} & GlaS~\cite{Warwick2015}  & DigestPath~\cite{Digest2019} \\ \hline
Positive & 10,000   & 4,000    & 4,521 & 5,954      \\ \hline
Negative & 10,000   & 4,000   & 4,143 & 6,000      \\ \hline
\end{tabular}
\label{datasets}
\end{table*}

\noindent\textbf{Validation of Unsupervised Domain Adaptation:} 
To validate the effectiveness of our proposed method, we compare our method against three widely investigated methods for unsupervised domain adaptation and the self-learning method without the GNN~\cite{graphconv2018} module, i.e., DANN~\cite{Ganin2016}, DeepJDOT~\cite{deepjdot}, and PixelDA~\cite{Bousmalis2017}. DANN~\cite{Ganin2016} and PixelDA~\cite{Bousmalis2017} are based on the theory of adversarial learning, while DeepJDOT~\cite{deepjdot} is based on the theory of optimal transport. `Our’s W/O GNN' is the variant of our method without the GNN~\cite{graphconv2018} module. Especially, DANN~\cite{Ganin2016} has been well implemented for the domain adaptation of histopathological images in previous works~\cite{Lafarge2017,Ren2018}. Table.~\ref{precision} records the classification precision of four compared methods, using two datasets as source domains respectively, i.e., CAMELYON16~\cite{Bejnordi2017} and DigestPath~\cite{Digest2019}. According to Table.~\ref{precision}, our proposed method can achieve superior performance in comparison with others, as well as consistently higher precisions across different datasets. These results are mainly benefited from the well-designed graph neural module. The newly presented loss function considers major variations by combining domain disparity, feature constraint~\cite{featureLoss2018} and graph module. Demonstrating our proposed method with the ability of tackling histopathological images from different clinical tasks.
\begin{table*}[t]
\centering
\normalsize
\caption{Performance comparison of four compared methods for unsupervised domain adaptation of histopathological images, under two datasets of source domains, i.e., CAMELYON16~\cite{Bejnordi2017} and DigestPath~\cite{Digest2019}.}
\resizebox{\textwidth}{!}{
\begin{tabular}{l|l|l|l|l|l|l}
\hline
 & \multicolumn{3}{l|}{CAMELYON16~\cite{Bejnordi2017}} & \multicolumn{3}{l}{DigestPath~\cite{Digest2019}} \\ \hline
 & BreakHis~\cite{Spanhol2016} & GlaS~\cite{Warwick2015} & DigestPath~\cite{Digest2019} & BreakHis~\cite{Spanhol2016} & GlaS~\cite{Warwick2015} & CAM16~\cite{Bejnordi2017} \\ \hline
DANN~\cite{Ganin2016} & 0.6803 & 0.5972 & 0.6437 & 0.6838 & 0.5145 & 0.5269 \\ \hline
DeepJDOT~\cite{deepjdot} & 0.7591 & \textbf{0.7796} & 0.6320 & 0.7024 & 0.5480 & 0.5945 \\ \hline
PixelDA~\cite{Bousmalis2017} & 0.7082 & 0.5573 & 0.6044 & 0.6532 & 0.6207 & 0.6332 \\ \hline
Our's W/O GNN & 0.6513 & 0.5375 & 0.6070 & 0.6692 & 0.6254 & 0.6244 \\ \hline
Our's W/ GNN & \textbf{0.7804} & 0.7542 & \textbf{0.7214} & \textbf{0.7548} & \textbf{0.6428} & \textbf{0.6416} \\ \hline
\end{tabular}}
\label{precision}
\end{table*}
\begin{figure}[t]
\centering
\includegraphics[width=.8\linewidth]{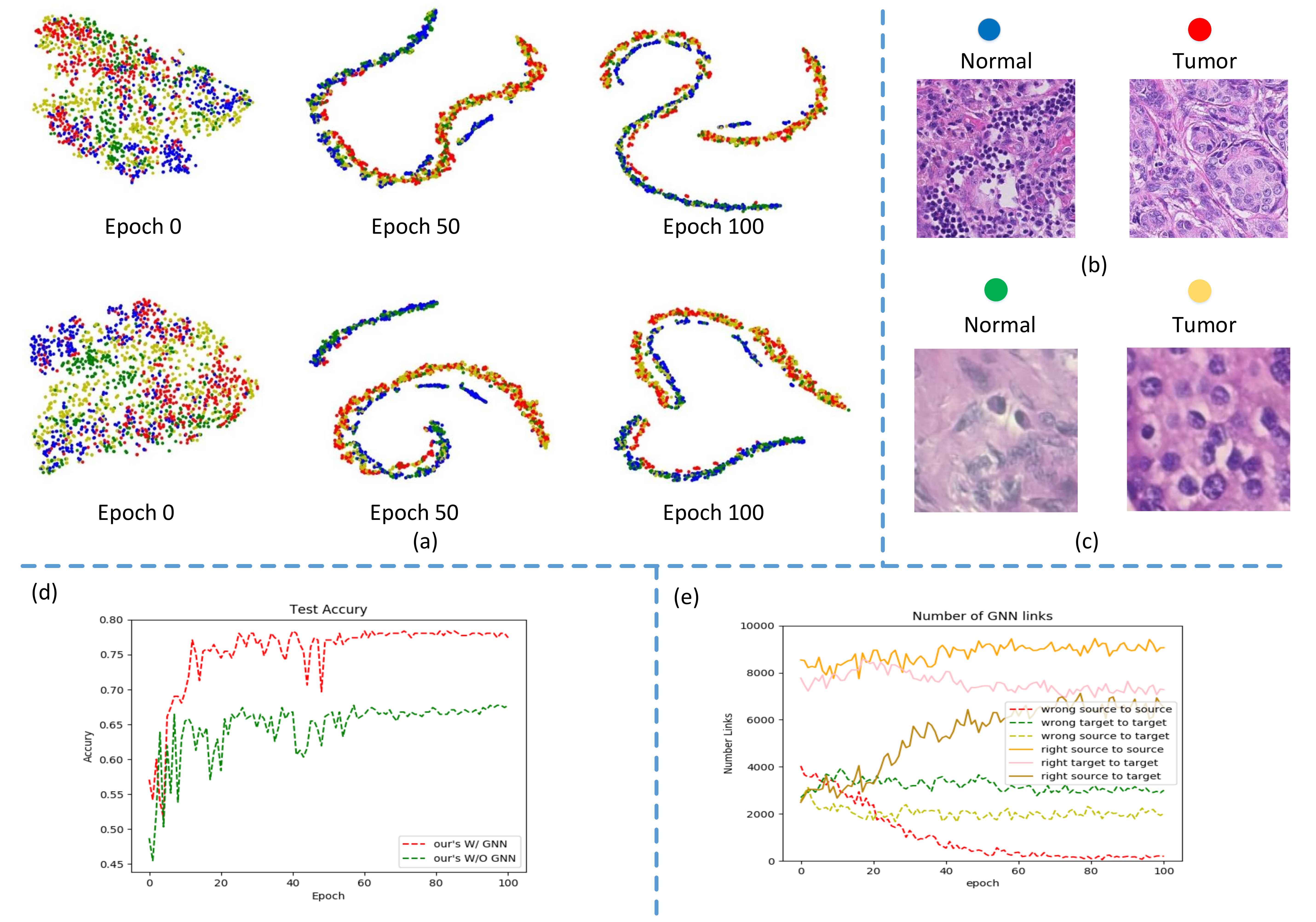}
\caption{(a) Feature distribution map of different training periods from 0 to 100, where the source image is extracted from CAMELYON16~\cite{Bejnordi2017} (indicating normal is blue and the tumor is red), and the target image is extracted from BreakHis~\cite{Spanhol2016} (indicating normal) It is green and the tumor is yellow; (b) and (c) are negative and positive pathological section data selected from the source domain (camelyon16) and the target domain (breakhis) respectively; (d) The classification accuracy of the target data with different epochs of training, the green line indicates that the GNN~\cite{graphconv2018} module is excluded, while the red indicates that the GNN~\cite{graphconv2018} module is included; (e) Number of statistical connections of features in the GNN~\cite{graphconv2018} module during the training process, 
where wrong indicates the connection of different labels, and right indicates the connection of the same label.}
\label{fig:gcn}
\end{figure}

\noindent\textbf{Evaluation of Graph Neural Network:} 
We further evaluate the effectiveness of the proposed graph neural module. We record the positions and connections of the feature updates of CAMELYON16~\cite{Bejnordi2017} (source domain) and BreakHis~\cite{Spanhol2016} (target domain) during different training epochs from 0 to 100. As shown in Fig.~\ref{fig:gcn}, blue and red circles indicate normal and tumor samples from the source domain, while green and yellow indicate normal and tumor samples from the target domain. According to Fig.~\ref{fig:gcn}(a), normal and tumor samples can be well classified once the GNN~\cite{graphconv2018} is adopted. In addition, samples from the source and target domains are more likely to be mixed together without domain differences. Fig.~\ref{fig:gcn}(b) and Fig.~\ref{fig:gcn}(c) show four images from source and target domains of normal and tumor respectively. Although these images demonstrate quite different appearance, they can be distinguished with same categories after the GNN~\cite{graphconv2018} is employed. Fig.~\ref{fig:gcn}(d) shows the comparison of classification accuracy between our proposed method with and without GNN~\cite{graphconv2018} module. According to Fig.~\ref{fig:gcn}(d), our proposed method with GNN~\cite{graphconv2018} module performs better than the method without GNN~\cite{graphconv2018} module, i.e., only through self-training in the final convergence stage. Fig.~\ref{fig:gcn}(e) shows the 
variations of the number of graph edges during the training stage. It is noticed that the number of correct connections can increase with the optimization process advances forward. 

\section{Conclusion}
In this paper, we propose a novel domain adaptation framework for the histopathological image analytics based on the graph neural networks. The developed GNN~\cite{graphconv2018} module can well differentiate cross-domain histopathological images showing large intra-class variations and small inter-class variations. The proposed loss function integrates the domain disparity, feature constraint and graph module to learn better feature representations for images in the target domain. Experimental results validate the effectiveness of the proposed framework on four public data sets with the comparison of state-of-the-arts.
%
%
%
%

\end{document}